\documentclass[letterpaper, 10 pt, conference]{ieeeconf}
\IEEEoverridecommandlockouts  %This command is only needed if you want to use the \thanks command

\usepackage{color}
\usepackage{array}

\usepackage{graphicx} % for pdf, bitmapped graphics files

\usepackage{mathptmx} % assumes new font selection scheme installed
\usepackage{times} % assumes new font selection scheme installed
\usepackage{amsmath} % assumes amsmath package installed
\usepackage{amssymb}  % as0sumes amsmath package installed
\usepackage{color}
\usepackage{bm}
\usepackage{rotating}
\usepackage{subfigure}
\usepackage{tabularx}
\usepackage{colortbl}
\usepackage{hhline}
\usepackage{multirow}
\usepackage{verbatim}
\usepackage{cite}
\usepackage{siunitx}
\usepackage[capitalise,noabbrev]{cleveref}
\usepackage{graphicx}

\usepackage{soul,xcolor}
\setstcolor{red}

\usepackage[acronym]{glossaries}
\newacronym{cnn}{CNN}{Convolutional Neural Network}

\newcommand{\tnet}{\texttt{touch2touch}}
\newcommand{\pnet}{\texttt{pix2pix}}
\newcommand{\resnet}{\texttt{ResNet18}}

\newcommand{\sat}{40000}

\begin{document}
	
\title{\LARGE \bf
	Touch-to-Touch Translation - Learning the Mapping Between Heterogeneous Tactile Sensing Technologies
 }

% Uncomment for IEEE template %%%%%%%%%%%%%%%%%%%%%%%%%%%%%%%%%%%%%%%%%%%%%%%%%%%%%%%

\author{Francesco Grella$^{1}$, Alessandro Albini$^{2*}$, Giorgio Cannata$^{1}$ and Perla Maiolino $^{2}$
\thanks{
$^{1}$ are with the DIBRIS, University of Genoa, IT.}
\thanks{$^{2}$ are with the Oxford Robotics Institute, University of Oxford, UK.}
\thanks{$^{*}$Corresponding author e-mail: alessandro@oxfordrobotics.institute}
\thanks{This work was supported by the SESTOSENSO (HORIZON EUROPE Research and Innovation Actions under GA number 101070310).}
% <-this % stops a space
}

\maketitle

%\linenumbers

\begin{abstract}
The use of data-driven techniques for tactile data processing and classification has recently increased. However, collecting tactile data is a time-expensive and sensor-specific procedure. Indeed, due to the lack of hardware standards in tactile sensing, data is required to be collected for each different sensor. 
%To avoid further data acquisition, 
This paper considers the problem of learning the mapping between two tactile sensor outputs with respect to the same physical stimulus - we refer to this problem as \textit{touch-to-touch translation}.
In this respect, we proposed two data-driven approaches to address this task and we compared their performance. The first one exploits a generative model developed for image-to-image translation and adapted for this context. The second one uses a ResNet model trained to perform a regression task.
%we propose \tnet{}, a generative model specifically trained to translate the output of one sensor to another. 
%
%
We validated both methods using two completely different tactile sensors - a camera-based, Digit \cite{Lambeta} and a capacitance-based, CySkin \cite{Schmitz}. In particular, we used Digit images to generate the corresponding  CySkin data. 
We trained the models on a set of tactile features that can be found in common larger objects and we performed the testing on a previously unseen set of data.
Experimental results show the possibility of translating Digit images into the CySkin output by preserving the contact shape and with an error of 15.18\% in the magnitude of the sensor responses. 
%In particular, \tnet{} is a generative model trained on 32 planar shapes in order to learn the mapping from one sensor to another. \textcolor{red}{togliere frase precedente?}

 %\textcolor{red}{parlare della loss sui taxels?}
%Moreover we adopt a metric based on 

%We deployed the model on a task of object recognition. 
% niente piu classificazione
\begin{comment}
In particular, we collected the dataset using Digit and we generate the corresponding outputs for CySkin. Results demonstrate that the object classifier trained with generated CySkin data has comparable performance when trained with real CySkin data, thus showing that this method can be used to reduce the effort required to collect data and to transfer the tactile knowledge between sensors.
\end{comment}
%
\end{abstract}

\section{Introduction}
\label{sec:intro}
Tactile sensing is fundamental to enable robots to physically interact with the environment \cite{Prete,Albini2020,Advait} and to perform complex manipulation tasks \cite{Lambeta,Li_acontrol,Kappassov}. 
In recent years, there has been a growing trend in employing data-driven algorithms to process tactile feedback \cite{LUO201754,Liu}.
However, the process of collecting a tactile dataset is inherently challenging and time-consuming, especially for classification tasks requiring multiple explorations of objects to capture their entire shape \cite{Pezzementi2011}.
To overcome the challenges of data collection, researchers have explored methods to artificially generate tactile data from camera images %In particular, the majority of the approaches exploit RGB-D cameras to generate tactile data  
\cite{Lee,Li_2019_CVPR,Patel,zhong_2022,Cai} by applying existing models for image-to-image translation to the tactile domain \cite{sym12101705}.
These generative models are trained on a large dataset of visuo-tactile data to learn the mapping between RGB-D data captured by cameras and the corresponding tactile response obtained through physical contact with the surface of interest. 
Despite the successful generation of tactile datasets (under certain conditions), unlike cameras, there is a lack of standardized hardware for tactile sensors which can significantly differ in terms of geometry, sensor arrangement, spatial resolution, transduction principles and mechanical properties \cite{Dahiya2010}. Consequently, models or algorithms developed for specific tactile hardware, cannot be directly applied to different sensors, often necessitating a new data collection procedure or model fine-tuning \cite{zhong_2022}.  The previously mentioned methods for data generation do not address this aspect.

\begin{figure*}[ht]
\centerline{\includegraphics[width=0.95\textwidth]{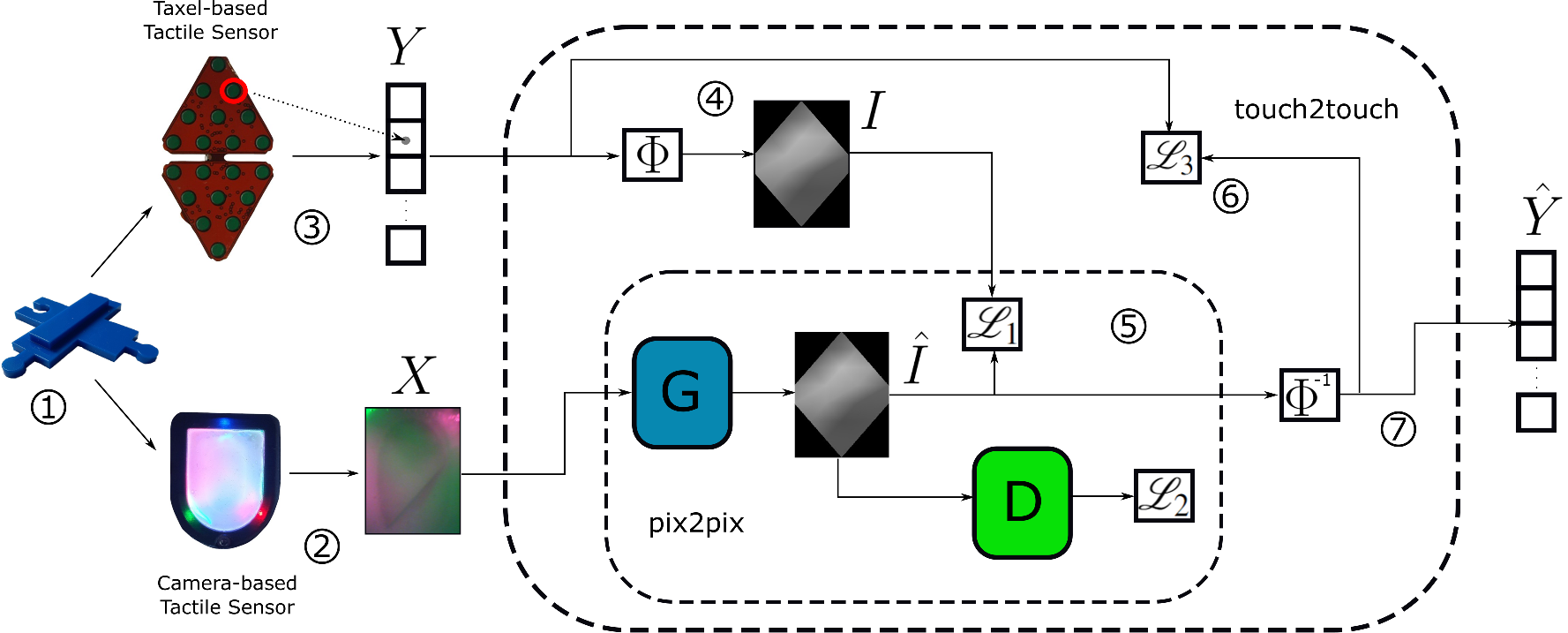}}
\caption{
	\tnet{} architecture and training pipeline. 
	(1) The network is trained by collecting paired tactile samples corresponding to the same physical stimulus;
        (2) Camera-based sensor and its output;
        %;
        (3) Taxel-based tactile sensor. The red circle highlight a single taxel which measurement is contained in the output array $Y$;
	(4) The output $Y$ is converted into an image $I$ and input to the \pnet{} model;
	(5) The \pnet{} model and its two loss functions as described in \cite{pix2pix}. G and D represent the generator and the discriminator;
	(6) The $\mathcal{L}_3$ loss added to properly stop the model training;
	(7) The output of \tnet{}, $\hat{Y}$ corresponding to $\hat{I}$ converted into an array.
	}
\label{img:overview}
\end{figure*}

In this paper, we take the initial step toward addressing this challenge by developing a method to transform data acquired with a specific tactile technology to the corresponding outputs that would be obtained using different hardware.

Our contribution is to present a method to learn the mapping $M:\{X\rightarrow Y\}$, with $X$ and $Y$ being the outputs of two distinct tactile sensors with respect to the same physical stimulus.
We refer to this task as \textit{touch-to-touch translation}. 
To the best of our knowledge, this paper represents the first attempt to address this problem. 
Specifically, we focus on translating images acquired with a camera-based tactile sensor \cite{Zhang} into the output of a system composed of distributed tactile elements (i.e. \textit{taxels}) \cite{Dahiya2010}. 

The proposed approach is tested using Digit \cite{Lambeta} and CySkin \cite{Schmitz} sensors. Therefore,  
%ated into CySkin responses (a capacitance-based sensor)\cite{Schmitz}. 
the model presented in this paper allows to artificially generate the CySkin output given a Digit tactile image as an input. 
It is important to remark that the two sensors are fundamentally different - 
Digit outputs images, while CySkin provides an array of measurements reflecting changes in capacitance between the two sensor layers.
These differences encompass spatial resolution, sensor geometry, spatial arrangement of the sensing elements, transduction mechanism and mechanical properties.
As a consequence, the responses of the two sensors with respect to the same tactile stimulus are substantially different, thus necessitating a model addressing these disparities. %translating between the two sensors needs to address these mismatches. 

We identified two possible strategies to address the touch-to-touch translation problem. The first one involves leveraging generative models developed for image-to-image translation and adapting them to our context. The second one consists of using Convolutional Neural Networks (CNNs) to perform a regression task, thus transforming the Digit input image into an array. In this respect, we perform a comparison to understand which approach is more suitable for this task. 

Furthermore, in order to generalize, the models should be trained on a wide variety of different objects such as in \cite{Li_2019_CVPR}, thus further complicating the data collection process. In this respect, we show that assuming a small contact area, the mapping between sensors can be learned on a set of features, representing tactile primitives such as line or step edges and corners that generally compose larger objects. Once the model is trained on this dataset it will be deployed on novel objects.

The remainder of the paper is organized as follows: \cref{sec:methodology} describes the two approaches and the training pipeline. %\textcolor{red}{and the object recognition task used to validate the proposed approach.} 
The experimental setup and the data collection procedure are explained in \cref{sec:dataset,sec:data_collection} respectively. Results and discussions are presented in \cref{sec:results}. Conclusion follows.

\section{Methodology}
\label{sec:methodology}

\begin{figure}[t]
	\centerline{\includegraphics[width=0.48\textwidth]{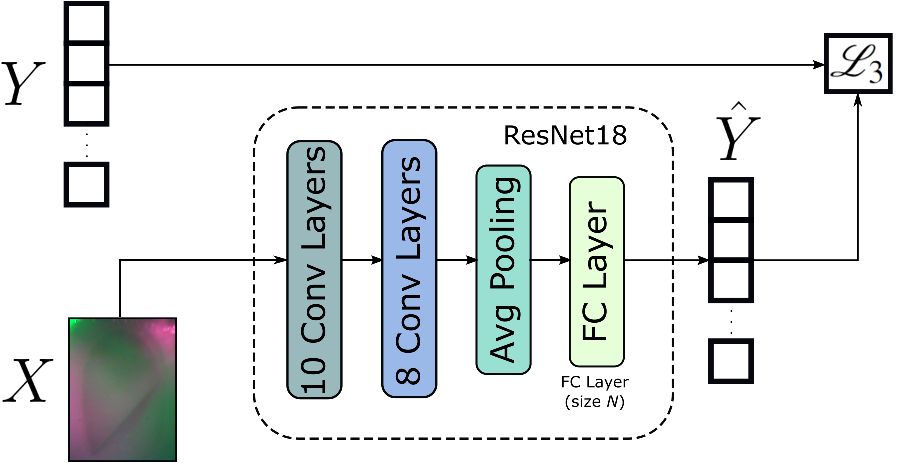}}
	\caption{
        The regression approach based on \resnet{} model. %The first 10 convolutional layers have been frozen during training, while the remaining sets of 9 convolutional layers have been retrained. 
        The softmax layer has been replaced with a fully connected layer of size $N$, corresponding to the number of taxels. The system is trained using the $\mathcal{L}_3$ loss. The shortcut connections of \resnet{} among layers have not been represented for simplicity.
	}
	\label{img:resnet}
\end{figure}%

\begin{figure*}[t]
	\centering
	\subfigure[]{\includegraphics[scale=0.3]{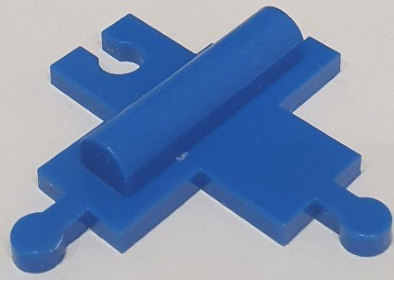}}
	\subfigure[]{\includegraphics[scale=0.3]{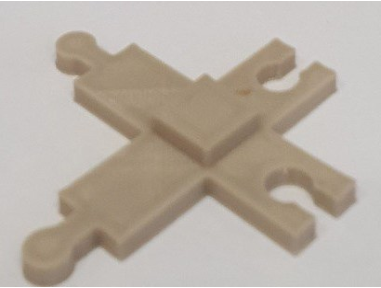}}
	\subfigure[]{\includegraphics[scale=0.31]{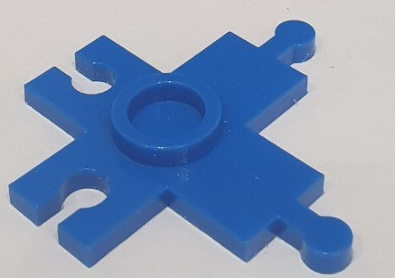}}
	\subfigure[]{\includegraphics[scale=0.3]{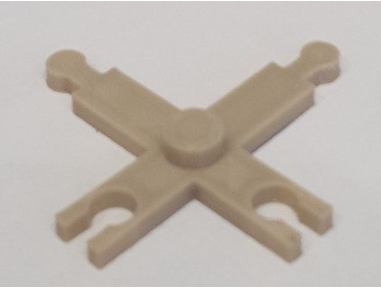}}
	\subfigure[]{\includegraphics[scale=0.29]{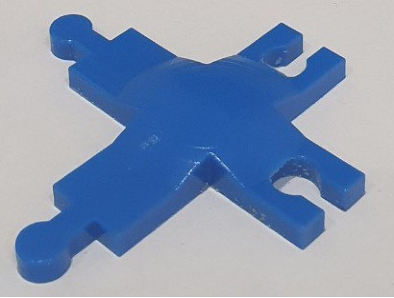}}
	\subfigure[]{\includegraphics[scale=0.3]{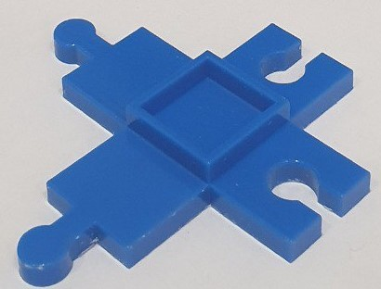}}
	\subfigure[]{\includegraphics[scale=0.29]{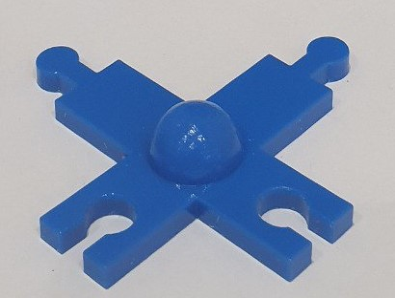}}
	\subfigure[]{\includegraphics[scale=0.3]{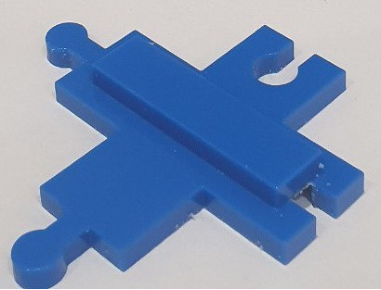}}
	\caption{ Tactile primitives used in this paper. The picture shows one for each of the 8 types. We 3D printed each one of them at 4 different scales, leading to a total of 32 primitives. (a) Line with smooth edges. (b) Square. (c) Empty Circle. (d) Circle. (e) Bump; (f) Empty square. (g) Hemisphere. (h) Line with sharp edges.}
	\label{img:features}
\end{figure*}
%\begin{comment}
\begin{figure*}[t]
	\centering
	\subfigure[]{\includegraphics[scale=0.16]{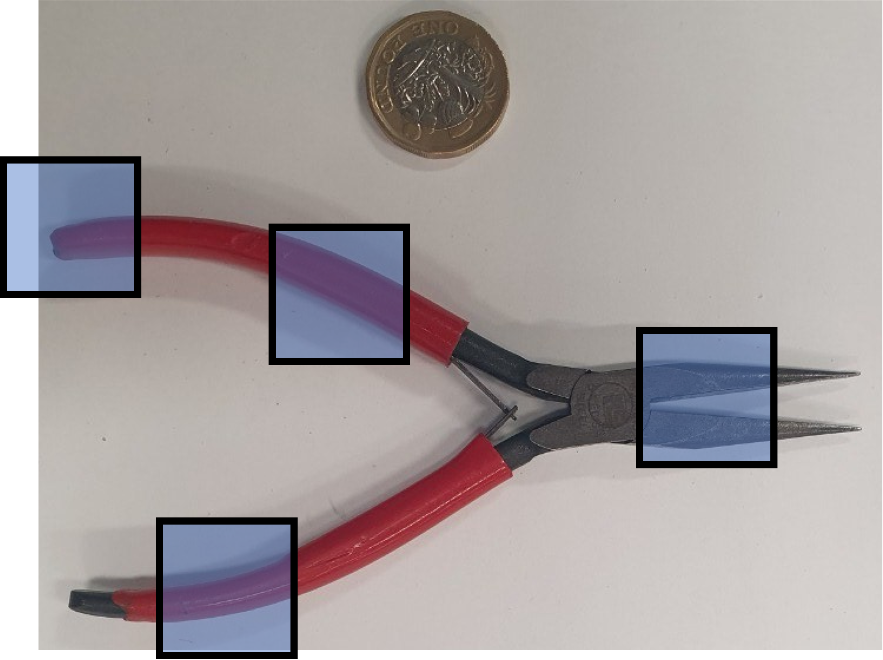}}
	\subfigure[]{\includegraphics[scale=0.16]{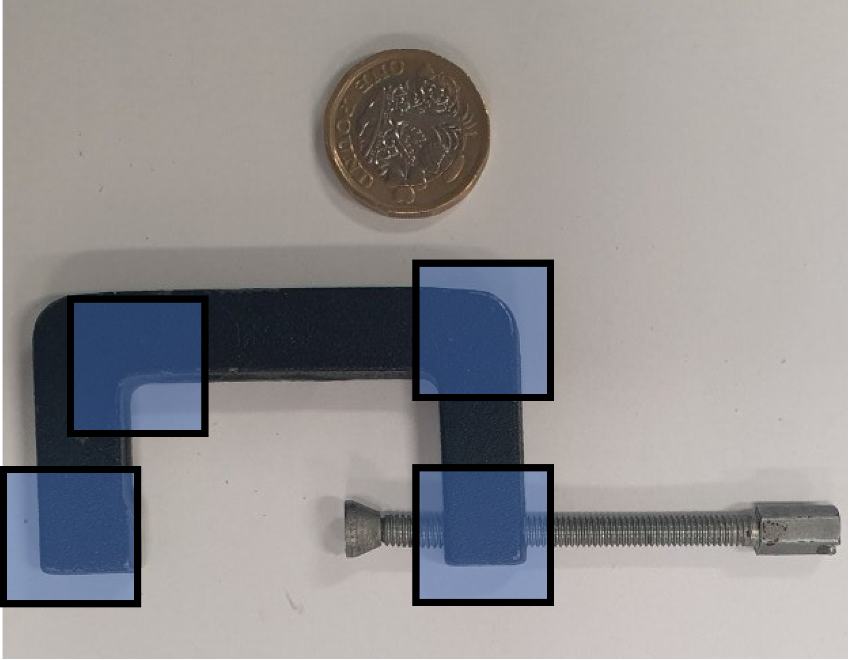}}
	\subfigure[]{\includegraphics[scale=0.16]{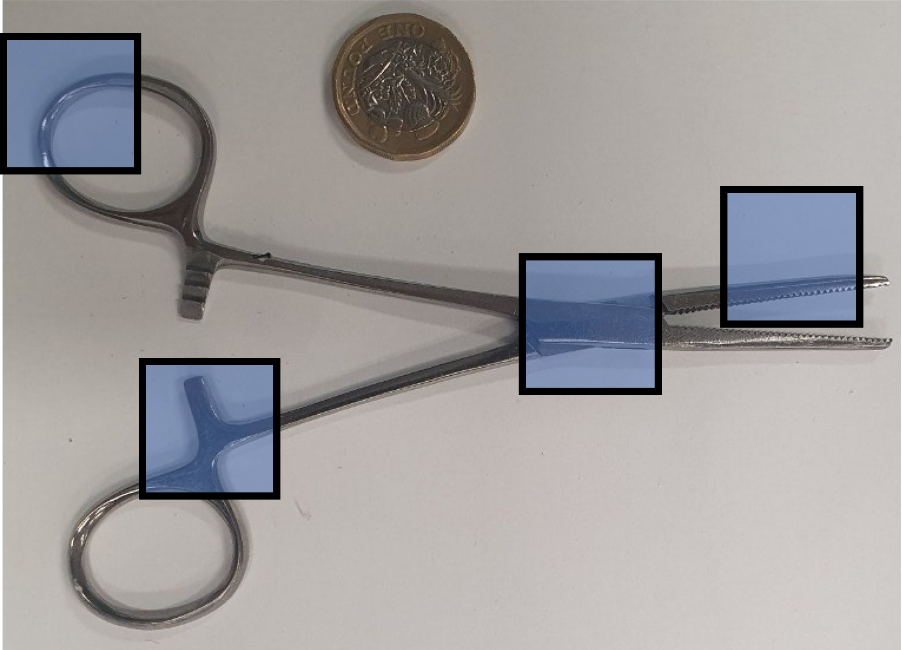}}
	\subfigure[]{\includegraphics[scale=0.16]{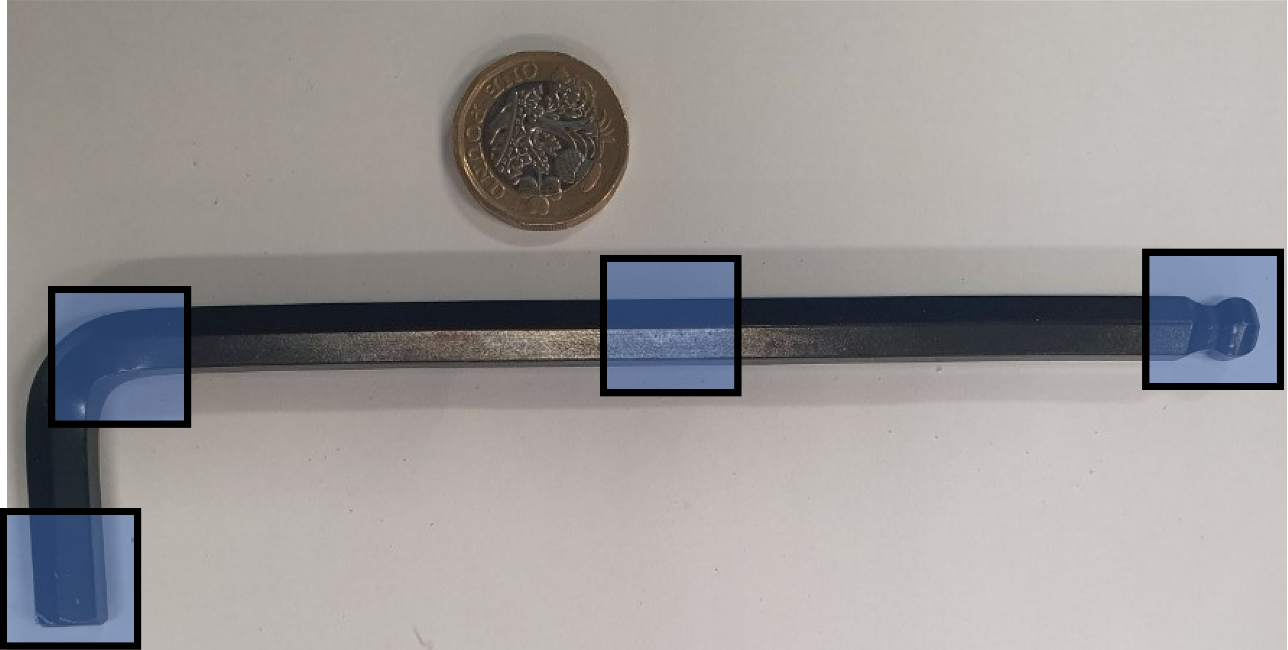}}
	\subfigure[]{\includegraphics[scale=0.16]{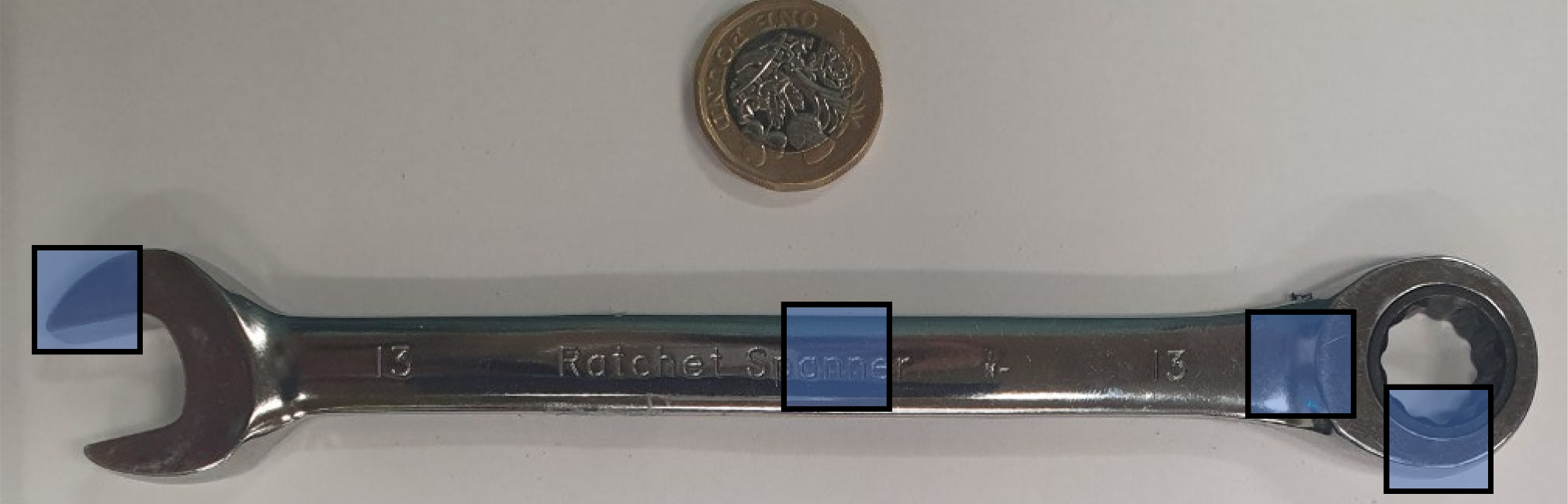}}
	\caption{ Objects used as a test set. The keypoints, corresponding to the sampling positions are marked with light blue squares on the objects. (a) Pliers. (b) Clamp. (c) Scissors. (d) Allen key. (e) Wrench. }
	\label{img:objects}
\end{figure*}

This Section describes the two architectures implemented to perform the translation and the relative training pipelines.
As previously discussed, we consider the mapping between a camera-based tactile sensor to a taxel-based system.
In the following, we denote with $X\in \mathbb{N}^{R\times C}$ the output of the camera-based sensor corresponding to an image of $R$ rows and $C$ columns. Similarly, the 
Furthermore, it is assumed $X$ and $Y$ to be \textit{paired} during the training phase - i.e. acquired in the same conditions in terms of contact location, sensor orientation and applied force.

The first approach consists of leveraging generative models used for image-to-image translation. In this case, we need to perform a set of transformations allowing for the network to handle data different from images. This is discussed in \cref{sec:gann}.
The second possible approach, described in \ref{sec:cnn}, consists in using a CNN to perform a regression task. In this case, data does not need to be transformed - the tactile image $X$ represents the input of the CNN, while the last fully connected layer corresponds to the output array of the taxel-based sensor. %This network can be trained to perform a regression task between the input image and the output array.

\subsection{Generative Model}
\label{sec:gann}

The proposed training pipeline is shown in \cref{img:overview}. The generative model, namely \tnet{}, is built on top of \pnet{} \cite{pix2pix} and designed to handle the differences between sensors.
As shown in \cref{img:overview}, we first transformed the array $Y$ (being the output of the taxel-based sensor) into a tactile image $I$. In this way, existing methods used for image-to-image translation, such as \pnet{}, can be applied. The output of the generator $\hat{I}$ is then converted back to an array $\hat{Y}$ matching the size of $Y$.

Although \pnet{} is designed to solve an image-to-image translation problem, we followed the same rationale as \cite{Cai} (where input arrays are converted into images representing spectrograms and vice versa) and we slightly modified the \pnet{} architecture to fit it to the considered task.

More in detail, as described in \cite{Albini2020}, we can convert the array of measurement $Y$ into a tactile image by performing an interpolation, thus defining $I = \phi(Y)$, where $\phi(\cdot)$ represents the operator performing a barycentric interpolation. 
This allows for framing our problem as an image-to-image translation task. 

is computed as $\hat{Y} = \phi^{\text{-}1} \left( \hat{I} \right) $, where $\hat{I}$ is the output image generated by \pnet{}. \\
This re-conversion step is necessary since to output the image $\hat{I}$ alone has a major disadvantage. 
Indeed, when considering a tactile sensor array, the tactile image is a representation of the sensor response that preserves the contact shape. While this can be useful for tasks such as object recognition \cite{Liu}, it may not generalize for other tasks, where the raw array measurement is preferred \cite{wangtactile}. 
Furthermore, there are different ways to interpolate the sensor responses with different spatial sampling steps that can depend on the specific task. 
As a consequence, considering $\hat{I}$ as the output will make the system task-specific.

In \cref{img:overview}, $\mathcal{L}_1$ and $\mathcal{L}_2$ are the two original losses of \pnet{}. Since from the standard \pnet{} losses it is not possible to reliably understand if the training has converged, we added a third loss $\mathcal{L}_3 = (Y - \hat{Y})^\intercal (Y - \hat{Y})$, not present in the \pnet{} original paper. This consists of an MSE loss. % between $Y$ and $\hat{Y}$. 
The smaller the MSE is, the more similar $Y$ and $\hat{Y}$ are. 
$\mathcal{L}_3$ does not directly affect the model's parameters. It is monitored during the training phase to apply an early stopping criterion and to properly decide when to stop the training. 
 
\subsection{CNN Regression Model}
\label{sec:cnn}

As previously mentioned, a CNN can be applied without requiring any data transformation. In this respect, we applied a \resnet{} model to our task \cite{resnet}. 
%The adopted model is a pre-trained ResNet18 with one input channel. 
In particular, we replace the original final fully connected layer of size 100 with a new one of size $N$ corresponding to the number of taxels (see \cref{img:resnet}).
The original training loss was changed from multi-class crossentropy to $\mathcal{L}_3$ (the same used for \tnet{}. 

It is worth noting that due to the presence of fully connected layers, \resnet{} can output a fixed-size array. 
Therefore, to apply this model to a sensor with a different number of tactile elements, its structure must be modified accordingly.
On the contrary, \tnet{} does not suffer from this limitation, since it does not contain fully connected layers and the data transformation added on top of \pnet{} can be applied to an array of arbitrary size.

\section{Experimental Setup}

\subsection{Dataset}
\label{sec:dataset}

The proposed model is trained on a dataset representing common features or \textit{tactile primitives}.
The dataset is partially inspired from \cite{Yashraj} and consists of 8 indenters resembling basic geometries that can be found in everyday objects and surfaces. 
The tactile features perceived by interacting with objects are usually line or step edges, ridges, sharp and smooth corners or large and small radii of curvatures of the local surface.
Assuming a fingertip-size contact area, our hypothesis is that a model trained on such primitives can be applied to more complex objects composed of similar features.

We 3D printed 8 types of features shown in  
\cref{img:features}. 
Furthermore, for each primitive, we considered four versions with different scales and edge thicknesses, thus leading to 32 tactile primitives. In particular, referring to \cref{img:features} the following parameters are changed: (a) width and radius; (b) side length; (c) radius and edge thickness; (d) radius; (e) height and curvature radius; (f) side length and edge thickness; (g) radius; (h) width.

In order to test the models on a set of previously unseen tactile data, we collected an additional dataset 
composed of 5 objects shown in \cref{img:objects}. Blue markers in \cref{img:objects} highlight the locations on the objects we defined to acquire tactile samples. 
The position of the sampling locations has been chosen to collect distinguished features from the objects, thus to test the models in different conditions.
Further information on the  data collection procedure is given in \cref{sec:data_collection_train,sec:data_collection_test}

\subsection{Tactile Sensors}

\begin{figure}[t]
  \centering
  \subfigure[]{\includegraphics[width=0.22\textwidth]{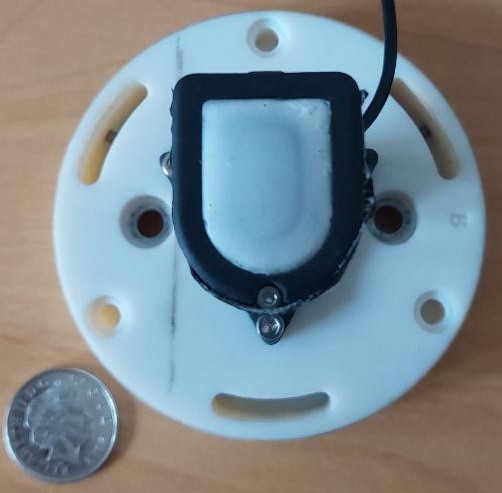} \label{img:digit}} \quad
  \subfigure[]{\includegraphics[width=0.217\textwidth]{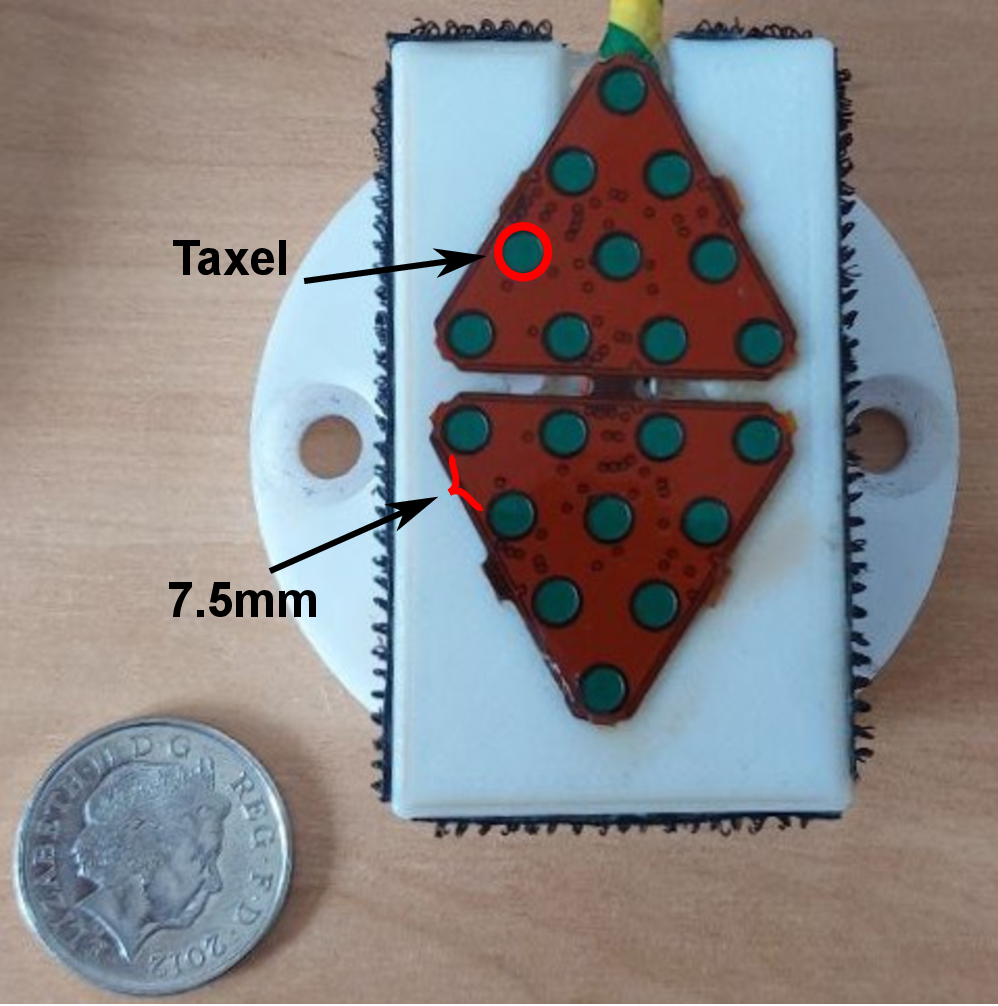} \label{img:cyskin}}
  \caption{The two tactile sensors used in this paper are integrated on a plastic mount that can be connected to the Panda Robot flange. (a) Digit provides an output RGB image of $320 \times 240$ pixels. The white soft layer on top is made on Solaris with Shore 15. (b) CySkin is a capacitive-based tactile sensor. The soft layer covering the sensor is made of Ecoflex Shore 10. In this picture, the soft layer has been removed to show the distribution of the tactile elements (the green circles). The pitch among them is \SI{7.5}{\milli\metre}. The sensor output corresponds to an array of 20 measurements.}
  \label{img:sensors}
\end{figure}

The two sensors used in this paper are Digit \cite{Lambeta} and CySkin \cite{Schmitz} (see \cref{img:sensors}). 
Digit is a camera-based sensors providing an RGB image of $320 \times 240$ pixels, while
CySkin is a capacitive-based tactile sensor which output consists of an array of 20 measurements corresponding to raw a 16-bit a-dimensional value related to the pressure value sensed by each taxel.
Although the theoretical full-scale of the sensor would be 65535, the sensor saturates at a value of \sat{}.
During training and evaluation, CySkin output is converted into a 1-channel tactile image of $396 \times 240$ (as in \cite{Albini2020}) using the operator $\Phi(\cdot)$ as reported in \cref{sec:gann}.

\section{Data Collection}

\label{sec:data_collection}

%%%%%%%%%%%%%%%%%%%%%%%%%% FIGURE

\subsection{Tactile Features Data Collection}
\label{sec:data_collection_train}

\begin{figure}[t]
	\centerline{\includegraphics[width=0.4\textwidth]{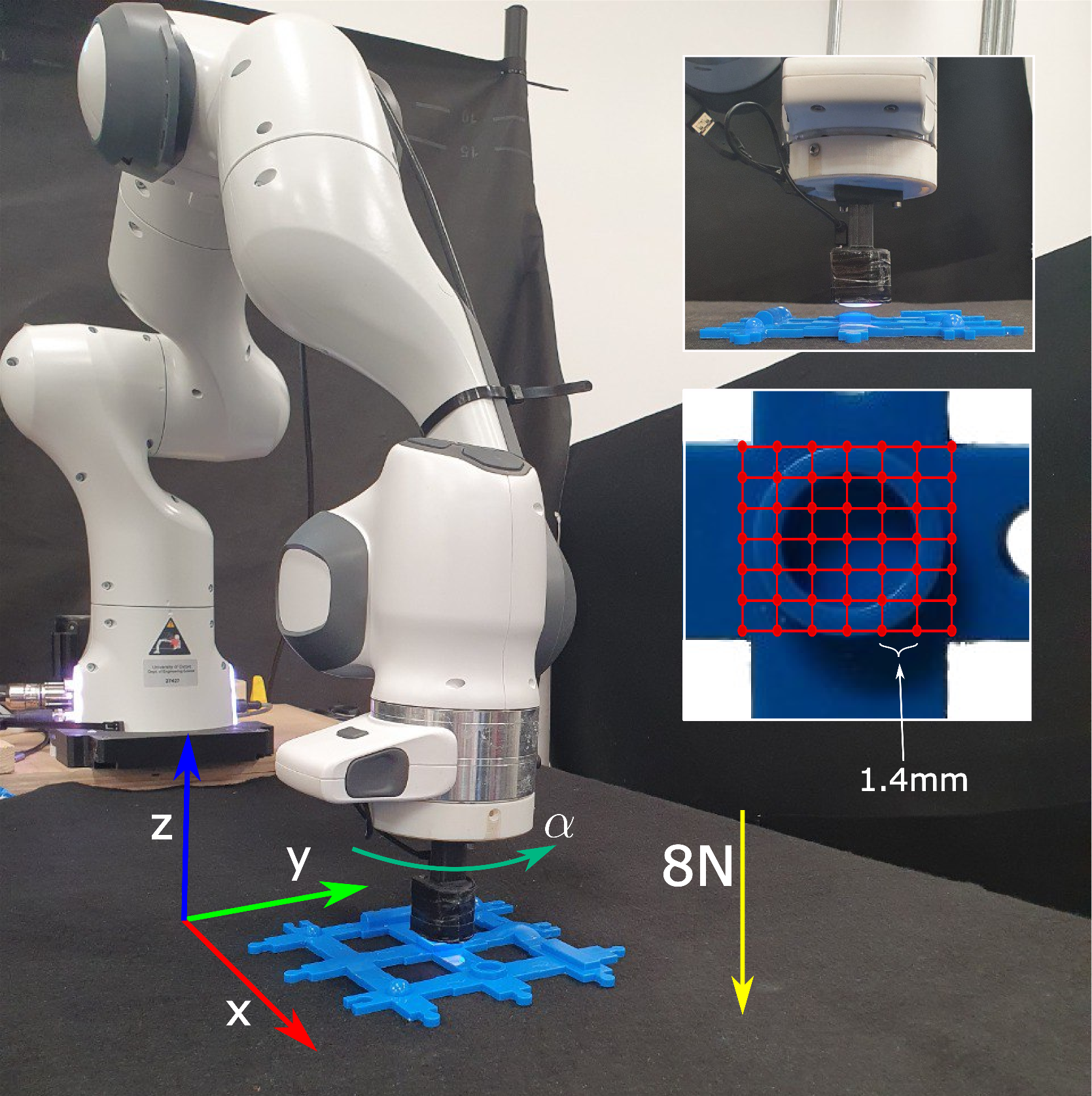}}
	\caption{
		Data collection procedure performed with both Digit and CySkin sensors. The panda robot is used to automate the process. A sampling grid is defined on top of each feature. The robot moves to these points and applies a controlled force along the $z$-axis.
	}
	\label{img:sampling}
\end{figure}

This data collection procedure must be performed once and it is only needed to learn the mapping between the two sensors. Indeed, after the model is trained, it can be deployed to generate artificial data from novel Digit tactile images.

We automated the data collection using a Panda Robot arm.
The Digit sensor is rigidly attached to the robot end-effector as shown in \cref{img:sampling}. Tactile features were connected in a mosaic manner and the robot was commanded to touch each tactile primitive. 
To train the models we collected a number of samples similar in order of magnitude to previous works dealing with visuo-tactile data generation such as \cite{Lee,Li_2019_CVPR}. 
In this respect, 
a planar sampling grid of 7x7 points is defined for each feature as shown in \cref{img:sampling} and it covers the whole shape of the tactile primitive.
Due to the different shapes and sizes, the grid resolution on the (x,y) plane changes depending on the type of feature. As an example, the resolution for the squared grid in  \cref{img:sampling} is \SI{1.4}{\milli\metre}.
For each point of the grid, we also define 12 different end-effector orientations around the plane $\alpha$, equally spaced and ranging between $0$ and $\frac{7}{4}\pi$. 
The robot is commanded to reach each point with the desired orientation and to apply a controlled force on the $z$-axis of \SI{8}{\newton}. 
Once the steady state is reached, we recorded 0.5 seconds of data and we computed the average to get a single tactile data sample. % is acquired. 
% leaving dynamic aspects to future extension of the work.
 
The data collection procedure was repeated for each point of the grid and each orientation, leading to a number of 588 samples for every single feature. 
Given the 32 different features, the total number of samples for Digit is 18816. %\lucio{update the number of total samples - UPDATED (virgola o punto in alto?)} \\
The same procedure was repeated with CySkin in the same experimental conditions thus acquiring additional 18816 samples paired (in terms of contact position and force) with the one collected with Digit.

It must be noted that the output of the sensor depends on the contact force applied on the surface. 
The generalization over multiple forces is a complex problem and it has never been addressed by past works dealing with tactile data generation \cite{Lee,Li_2019_CVPR,Patel,zhong_2022}. Therefore, to address with this aspect requires a further deep analysis which is out of the scope of this paper.
We leave these analyses to a future extension of the work.

The dataset was split into two subsets for training, and validation. As previously explained, we printed 4 different versions of the tactile primitive. To make sure to validate the performance of the model on previously unseen data, we split it such that the subsets contains different versions of the features with a proportion of 3:1 for training and validation respectively.

\subsection{Objects Data Collection}
\label{sec:data_collection_test}
% Objects
The data collection procedure for testing the proposed approaches on previously unseen data
is similar to the one adopted for the basic features. 
We collected tactile data of the 5 objects shown in \cref{img:objects}. We sampled the single object by defining 4 different keypoints on it, which are highlighted in \cref{img:objects}. A sampling grid was then defined in the centre of each keypoint.
For testing purposes, we need a lower number of samples compared to the training case. Therefore, we used a sampling grid of 3x3 points with a resolution of  \SI{1.5}{\milli\metre}. For each grid we collected 63 tactile samples, reaching a total of 1260 samples for the whole objects set. This procedure was performed for both Digit and CySkin sensors.

\section{Results and Discussion}
\label{sec:results}

\begin{figure*}[t]
	\centerline{\includegraphics[width=0.97\textwidth]{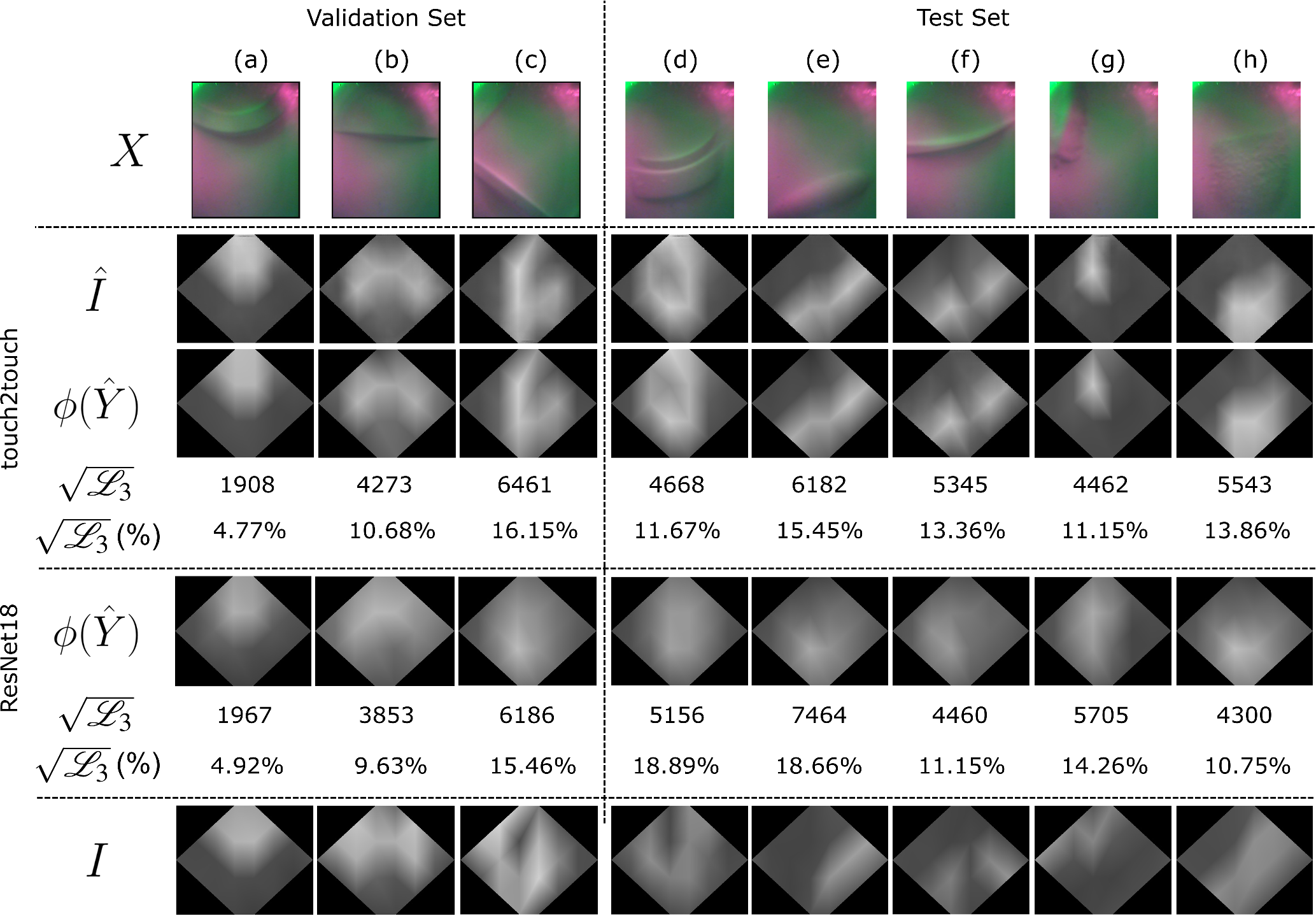}}
	\caption{
		Output of \tnet{} and \resnet{} in form of images. (a)-(c) represent tactile data belonging to the validation set (tactile primitives) while (d)-(e) belong to the test set of novel objects. 
		The first row represents the Digit output; 
		$\hat{I}$ is tactile image in output from \pnet{} (see \cref{img:overview}); 
		$\phi(\hat{Y})$ corresponds to the barycentric interpolation applied to the output $\hat{Y}$; 
		Figure also reports $\sqrt{\mathcal{L}_3}$ and the error in percentage with respect to the CySkin fullscale.
		The last row represents the tactile image generated from real CySkin data $Y$. 
	}
	\label{img:net_out}
\end{figure*}

Both models were trained on the set of tactile primitives described in \cref{sec:data_collection_train}. %The validation set was used to property stop the training. 
The hyper-parameters used to train \tnet{} are the same used for \pnet{} and they can be found in \cite{pix2pix}.
The only exception is for the batch size that we set to 32. 
The \pnet{} model requires tactile images from Digit and CySkin to have the same size. We added additional resize layers to \tnet{} to perform this operation online. We kept the same input size used in the \pnet{} paper, thus, the images were resized to $256 \times 256$ pixels. 
Since the CySkin tactile image is 1-channel, we converted the Digit output to a grayscale image.
For what concerns the \resnet{} model we used a fixed learning rate of 0.001 and a batch size of 8.

For both models, the training was stopped by monitoring $\mathcal{L}_3$ on the validation set (see \cref{sec:data_collection_train}) and using the early stop criterion \cite{Goodfellow}.
The models were then applied to the test set of objects. Their performance can be qualitatively assessed by evaluating $\sqrt{\mathcal{L}_3}$ (i.e. the RMSE between $Y$ and $\hat{Y}$) on the whole test set.
We obtained $\sqrt{\mathcal{L}_3} = 6072$ and $\sqrt{\mathcal{L}_3} = 4867$ for \tnet{} and \resnet{} respectively. 
Considering that the CySkin raw output saturates at \sat{}, this corresponds to a percentage error of 15.18\% for \tnet{} and of 12.17\% for \resnet{} with respect to the fullscale. 
% Add table
\begin{table}[t]
	\centering
	\caption{$\sqrt{\mathcal{L}_3}$ reported for both models and averaged on the whole sets.}
	\label{tb:results}
\begin{tabular}{cc|c|c|}
	\cline{3-4}
	\multicolumn{1}{l}{}                                                                                      & \multicolumn{1}{l|}{} & \tnet{} & \resnet{} \\ \hline
	%\multicolumn{1}{|c|}{\multirow{3}{*}{\textbf{\begin{tabular}[c]{@{}c@{}}Validation \\ Set\end{tabular}}}} & $\sqrt{\mathcal{L}_3}$  & 4007 & 3931 \\
     \multicolumn{1}{|c|}{\multirow{2}{*}{\textbf{\begin{tabular}[c]{@{}c@{}}Validation \\ Set\end{tabular}}}} & $\sqrt{\mathcal{L}_3}$  & 4007 & 3931 \\
	\multicolumn{1}{|c|}{}                                                                                    & $\sqrt{\mathcal{L}_3}$ (\%) & 10.01      & 9.83       % \\
	%\multicolumn{1}{|c|}{}                                                                                    & SSIM                  & 0.99      & 0.98        
    \\ \hline
	%\multicolumn{1}{|c|}{\multirow{3}{*}{\textbf{\begin{tabular}[c]{@{}c@{}}Test \\ Set\end{tabular}}}}       & $\sqrt{\mathcal{L}_3}$                 & 6072      & 4867        \\
    \multicolumn{1}{|c|}{\multirow{2}{*}{\textbf{\begin{tabular}[c]{@{}c@{}}Test \\ Set\end{tabular}}}}       & $\sqrt{\mathcal{L}_3}$                 & 6072      & 4867        \\
	\multicolumn{1}{|c|}{}                                                                                    & $\sqrt{\mathcal{L}_3}$ (\%)            & 15.18      & 12.17       % \\
	%\multicolumn{1}{|c|}{}                                                                                    & SSIM                  & 0.92      & 0.96        
    \\ \hline
\end{tabular}
\end{table}
%%%%%%%%%%%
%
%

It can be seen that the models perform slightly worst on the test set. This is reasonable - the validation set contains variations (in terms of size) of previously seen tactile features while the test set is composed of novel objects (see \cref{img:objects}).
Although features in the test set could be similar to the ones used for training in terms of overall shape, they still differ for small details: they could have different curvatures, corners and textures of the surface.
Due to its high resolution, Digit can capture all these differences not existing in the training set.

From \cref{tb:results} is clear that \resnet{} performs better than \tnet{} in terms of $\mathcal{L}_3$ loss. 
However, we conducted a further analysis by considering the generated outputs in terms of tactile images.
In this respect, we compared $I$, i.e. the tactile image corresponding to the real CySkin output $Y$, with $\phi(\hat{Y})$ being the image created by converting the generated output $\hat{Y}$.

\cref{img:net_out} shows data in form of images along with the corresponding $\sqrt{\mathcal{L}_3}$. % for a number of samples belonging to the test set. 
In particular, Figures \ref{img:net_out}(a)-(c) refer to data belonging to the validation set, while Figures \ref{img:net_out}(d)-(h) are related to the test set of novel objects.
The first and last rows in \cref{img:net_out} show the Digit input $X$ and the tactile image $I$. %obtained by applying the barycentric interpolation to real CySkin data. 
The second row corresponds to the output of \pnet{} $\hat{I}$ (see \cref{img:overview}).
The Figure also reports $\phi(\hat{Y})$, $\sqrt{\mathcal{L}_3}$ and the error in percentage for both models. 
It is important to remark that in the case of \tnet{} $\hat{I} \neq \phi(\hat{Y})$. As visible from the Figures the output of the generator $\hat{I}$ is slightly blurred compared to the real tactile image $I$. 
This means that the \pnet{} can generate images preserving the contact shape but cannot exactly approximate the function $\phi(\cdot)$ performing a barycentric interpolation. This aspect is of low interest in our application since the goal is to generate an array of measurements.

Images were first compared using the SSIM index as proposed in \cite{Lee}. In this respect, we obtained an SSIM of 0.96 and 0.95 for \tnet{} and \resnet{} respectively. However, despite the two values being nearly similar, SSIM is not always reliable for quantifying the quality of the generated images \cite{nilsson2020understanding,7351345}. Indeed, it can be seen from \cref{img:net_out} that \tnet{} produces an output much more similar to the original than \resnet{}. 

A qualitative analysis of \cref{img:net_out} shows that
\tnet{} is able to 
generate an output which better preserves the contact shape.  %\rev{The SSIM index cannot capture these difference in the overall shape of the image \cite{aa,aa}}.
Looking at the $\phi(\hat{Y})$ rows in the Figure and comparing them with the ground-truth $I$ it can be seen how \tnet{} produces sharper contact distribution closer to the real one, while the output of  \resnet{} is blurred and in most of the cases does not match real contact shape.
We argue that this is due to the training pipeline based on \pnet{}. Indeed, the conversion into images  (see \cref{img:overview}) allows for the network to learn spatial relations among CySkin tactile elements.
On the contrary, spatial information is missing during the training of \resnet{}, causing the network to learn how to minimize $\mathcal{L}_3$ but producing an array $\hat{Y}$ not matching the original contact distribution. 
Even when considering the validation set (Figures \ref{img:net_out}(a)-(c)), consisting of variations of features used during training, \resnet{} can hardly reconstruct the correct shape. \newline
The last five columns represent samples related to the objects in \cref{img:objects}. 
Figures \ref{img:net_out}(d)-(f) show three samples generated by \tnet{} where the contact distribution is generally preserved and the major difference is in the magnitude. 
Figures \ref{img:net_out}(g)-(h) represent instead poorly generated tactile samples where the shape was not well preserved. It is worth noting that even in these last two cases, \tnet{} still performs better than \resnet{}.

Therefore, we conclude that in this task of touch-to-touch translation, a generative-based approach is a better choice compared to methods performing a regression, since they allow for preserving spatial information. 
It must also be noted that inputs $X$ in \cref{img:net_out} have not previously been seen by the model during training. Indeed, as explained in \cref{sec:data_collection_train}, the validation set was created to contain unseen variations of the tactile primitive. The test set instead contains data collected from different objects. Therefore, since the model is able to reconstruct the local contact shape we can assess that \tnet{} can be effectively trained on tactile primitives to generalize on more complex objects.
These findings are useful to drive future research on this topic. For example, an additional loss function could be added to further enforce the reconstruction of the overall contact shape, thus improving the performance on novel objects. This aspect will be investigated in a future extension of the paper.

\section{Conclusion}

Tactile data are hard to collect and technology-specific.
Therefore, when changing sensor, a new data collection procedure  is often required.
In this paper, we tackle this problem by proposing two data-driven approaches to learn the mapping between two sensor outputs. The first solution is based on generative models for image-to-image translation. We adapted the existing \pnet{} architecture to work with tactile data. The second solution is based on a regression approach exploiting a \resnet{} model.
Both models were trained on a dataset of tactile primitives resembling features that can be found in more complex objects. The approaches were then tested on a different dataset composed of previously unseen objects. 
Experimental results show that the solution exploiting generative models for image-to-image translation is better suited for this task. Indeed, the proposed training pipeline, embedding the conversion of data into tactile images, allows for the network to preserve the contact shape.

There are still open questions. For example, in this work, we only consider the generation from a high-resolution sensor to a low-resolution one. In the opposite case, the low-resolution sensors cannot capture fine details, therefore, the high-resolution data would be harder to generate. An analysis focusing on this aspect, as well as on the generalization to different contact forces, will be carried out in a future extension of the paper. 
It must also be noted that the proposed method is based on the assumption that the contact area is small, such that the tactile features in \cref{sec:dataset} can be retrieved into more complex objects. Therefore, what is proposed in this paper is applicable to fingertip-size sensors (which are the most widespread among the robotics community).

{\small
	\bibliographystyle{IEEEtran}
	\bibliography{ref}

\begin{thebibliography}{10}
\providecommand{\url}[1]{#1}
\csname url@samestyle\endcsname
\providecommand{\newblock}{\relax}
\providecommand{\bibinfo}[2]{#2}
\providecommand{\BIBentrySTDinterwordspacing}{\spaceskip=0pt\relax}
\providecommand{\BIBentryALTinterwordstretchfactor}{4}
\providecommand{\BIBentryALTinterwordspacing}{\spaceskip=\fontdimen2\font plus
\BIBentryALTinterwordstretchfactor\fontdimen3\font minus
  \fontdimen4\font\relax}
\providecommand{\BIBforeignlanguage}[2]{{%
\expandafter\ifx\csname l@#1\endcsname\relax
\typeout{** WARNING: IEEEtran.bst: No hyphenation pattern has been}%
\typeout{** loaded for the language `#1'. Using the pattern for}%
\typeout{** the default language instead.}%
\else
\language=\csname l@#1\endcsname
\fi
#2}}
\providecommand{\BIBdecl}{\relax}
\BIBdecl

\bibitem{Lambeta}
M.~Lambeta, P.-W. Chou, S.~Tian, B.~Yang, B.~Maloon, V.~R. Most, D.~Stroud,
  R.~Santos, A.~Byagowi, G.~Kammerer, D.~Jayaraman, and R.~Calandra, ``Digit: A
  novel design for a low-cost compact high-resolution tactile sensor with
  application to in-hand manipulation,'' \emph{IEEE Robotics and Automation
  Letters}, vol.~5, no.~3, pp. 3838--3845, 2020.

\bibitem{Schmitz}
A.~Schmitz, P.~Maiolino, M.~Maggiali, L.~Natale, G.~Cannata, and G.~Metta,
  ``Methods and technologies for the implementation of large-scale robot
  tactile sensors,'' \emph{IEEE Transactions on Robotics}, vol.~27, no.~3, pp.
  389--400, 2011.

\bibitem{Prete}
A.~Del~Prete, F.~Nori, G.~Metta, and L.~Natale, ``Control of contact forces:
  The role of tactile feedback for contact localization,'' in \emph{2012
  IEEE/RSJ International Conference on Intelligent Robots and Systems}, 2012,
  pp. 4048--4053.

\bibitem{Albini2020}
\BIBentryALTinterwordspacing
A.~Albini and G.~Cannata, ``Pressure distribution classification and
  segmentation of human hands in contact with the robot body:,''
  \emph{https://doi.org/10.1177/0278364920907688}, vol.~39, pp. 668--687, 3
  2020. [Online]. Available:
  \url{https://journals.sagepub.com/doi/full/10.1177/0278364920907688}
\BIBentrySTDinterwordspacing

\bibitem{Advait}
\BIBentryALTinterwordspacing
A.~Jain, M.~D. Killpack, A.~Edsinger, and C.~C. Kemp, ``Reaching in clutter
  with whole-arm tactile sensing,'' \emph{The International Journal of Robotics
  Research}, vol.~32, no.~4, pp. 458--482, 2013. [Online]. Available:
  \url{https://doi.org/10.1177/0278364912471865}
\BIBentrySTDinterwordspacing

\bibitem{Li_acontrol}
Q.~Li, C.~Schürmann, R.~Haschke, and H.~Ritter, ``A control framework for
  tactile servoing.''

\bibitem{Kappassov}
\BIBentryALTinterwordspacing
Z.~Kappassov, J.-A. Corrales, and V.~Perdereau, ``Touch driven controller and
  tactile features for physical interactions,'' \emph{Robotics and Autonomous
  Systems}, vol. 123, p. 103332, 2020. [Online]. Available:
  \url{https://www.sciencedirect.com/science/article/pii/S0921889019300697}
\BIBentrySTDinterwordspacing

\bibitem{LUO201754}
\BIBentryALTinterwordspacing
S.~Luo, J.~Bimbo, R.~Dahiya, and H.~Liu, ``Robotic tactile perception of object
  properties: A review,'' \emph{Mechatronics}, vol.~48, pp. 54--67, 2017.
  [Online]. Available:
  \url{https://www.sciencedirect.com/science/article/pii/S0957415817301575}
\BIBentrySTDinterwordspacing

\bibitem{Liu}
\BIBentryALTinterwordspacing
H.~Liu, Y.~Wu, F.~Sun, and D.~Guo, ``Recent progress on tactile object
  recognition,'' \emph{International Journal of Advanced Robotic Systems},
  vol.~14, no.~4, p. 1729881417717056, 2017. [Online]. Available:
  \url{https://doi.org/10.1177/1729881417717056}
\BIBentrySTDinterwordspacing

\bibitem{Pezzementi2011}
Z.~Pezzementi, E.~Plaku, C.~Reyda, and G.~D. Hager, ``Tactile-object
  recognition from appearance information,'' \emph{IEEE Transactions on
  Robotics}, vol.~27, pp. 473--487, 6 2011.

\bibitem{Lee}
J.-T. Lee, D.~Bollegala, and S.~Luo, ``“touching to see” and “seeing to
  feel”: Robotic cross-modal sensory data generation for visual-tactile
  perception,'' in \emph{2019 International Conference on Robotics and
  Automation (ICRA)}, 2019, pp. 4276--4282.

\bibitem{Li_2019_CVPR}
Y.~Li, J.-Y. Zhu, R.~Tedrake, and A.~Torralba, ``Connecting touch and vision
  via cross-modal prediction,'' in \emph{Proceedings of the IEEE/CVF Conference
  on Computer Vision and Pattern Recognition (CVPR)}, June 2019.

\bibitem{Patel}
K.~Patel, S.~Iba, and N.~Jamali, ``Deep tactile experience: Estimating tactile
  sensor output from depth sensor data,'' in \emph{2020 IEEE/RSJ International
  Conference on Intelligent Robots and Systems (IROS)}, 2020, pp. 9846--9853.

\bibitem{zhong_2022}
\BIBentryALTinterwordspacing
S.~Zhong, A.~Albini, O.~P. Jones, P.~Maiolino, and I.~Posner, ``Touching a
  ne{RF}: Leveraging neural radiance fields for tactile sensory data
  generation,'' in \emph{6th Annual Conference on Robot Learning}, 2022.
  [Online]. Available: \url{https://openreview.net/forum?id=No3mbanRlZJ}
\BIBentrySTDinterwordspacing

\bibitem{Cai}
S.~Cai, K.~Zhu, Y.~Ban, and T.~Narumi, ``Visual-tactile cross-modal data
  generation using residue-fusion gan with feature-matching and perceptual
  losses,'' \emph{IEEE Robotics and Automation Letters}, vol.~6, no.~4, pp.
  7525--7532, 2021.

\bibitem{sym12101705}
\BIBentryALTinterwordspacing
A.~Alotaibi, ``Deep generative adversarial networks for image-to-image
  translation: A review,'' \emph{Symmetry}, vol.~12, no.~10, 2020. [Online].
  Available: \url{https://www.mdpi.com/2073-8994/12/10/1705}
\BIBentrySTDinterwordspacing

\bibitem{Dahiya2010}
\BIBentryALTinterwordspacing
R.~S. Dahiya, G.~Metta, M.~Valle, and G.~Sandini, ``Tactile sensing-from humans
  to humanoids,'' \emph{IEEE Transactions on Robotics}, vol.~26, pp. 1--20, 2
  2010. [Online]. Available:
  \url{https://ieeexplore.ieee.org/stamp/stamp.jsp?arnumber=5339133}
\BIBentrySTDinterwordspacing

\bibitem{pix2pix}
P.~Isola, J.-Y. Zhu, T.~Zhou, and A.~A. Efros, ``Image-to-image translation
  with conditional adversarial networks,'' in \emph{2017 IEEE Conference on
  Computer Vision and Pattern Recognition (CVPR)}, 2017, pp. 5967--5976.

\bibitem{Zhang}
S.~Zhang, Z.~Chen, Y.~Gao, W.~Wan, J.~Shan, H.~Xue, F.~Sun, Y.~Yang, and
  B.~Fang, ``Hardware technology of vision-based tactile sensor: A review,''
  \emph{IEEE Sensors Journal}, vol.~22, no.~22, pp. 21\,410--21\,427, 2022.

\bibitem{wangtactile}
S.-a. WANG, A.~Albini, P.~Maiolino, F.~Mastrogiovanni, and G.~Cannata,
  ``Tactile based fabric classification via robotic sliding,'' \emph{Frontiers
  in Neurorobotics}, p.~10.

\bibitem{resnet}
K.~He, X.~Zhang, S.~Ren, and J.~Sun, ``Deep residual learning for image
  recognition,'' 06 2016, pp. 770--778.

\bibitem{Yashraj}
\BIBentryALTinterwordspacing
Y.~S. Narang, B.~Sundaralingam, K.~V. Wyk, A.~Mousavian, and D.~Fox,
  ``Interpreting and predicting tactile signals for the syntouch biotac,''
  \emph{The International Journal of Robotics Research}, vol.~40, no. 12-14,
  pp. 1467--1487, 2021. [Online]. Available:
  \url{https://doi.org/10.1177/02783649211047634}
\BIBentrySTDinterwordspacing

\bibitem{Goodfellow}
I.~Goodfellow, Y.~Bengio, and A.~Courville, \emph{Deep Learning}.\hskip 1em
  plus 0.5em minus 0.4em\relax MIT Press, 2016,
  \url{http://www.deeplearningbook.org}.

\bibitem{nilsson2020understanding}
J.~Nilsson and T.~Akenine-M{\"o}ller, ``Understanding ssim,'' \emph{arXiv
  preprint arXiv:2006.13846}, 2020.

\bibitem{7351345}
J.-F. Pambrun and R.~Noumeir, ``Limitations of the ssim quality metric in the
  context of diagnostic imaging,'' in \emph{2015 IEEE International Conference
  on Image Processing (ICIP)}, 2015, pp. 2960--2963.

\end{thebibliography}
}

\end{document}